\documentclass{article}
\usepackage{arxiv}
\usepackage[utf8]{inputenc} % allow utf-8 input
\usepackage[T1]{fontenc}    % use 8-bit T1 fonts
\usepackage{hyperref}       % hyperlinks
\usepackage{url}            % simple URL typesetting
\usepackage{booktabs}       % professional-quality tables
\usepackage{amsfonts}       % blackboard math symbols
\usepackage{nicefrac}       % compact symbols for 1/2, etc.
\usepackage{microtype}      % microtypography
\usepackage{lipsum}		% Can be removed after putting your text content
\usepackage{graphicx}
\usepackage{doi}
\usepackage{geometry}
\usepackage{siunitx}
\usepackage{amsmath}
\usepackage[backend=biber, style=apa]{biblatex}
\usepackage{lscape}
\usepackage{array,multirow,graphicx}

\title{Learning COVID-19 Regional Transmission Using Universal Differential Equations in a SIR model}
\author{Adrián Rojas-Campos, Lukas Stelz and Pascal Nieters}
\date{October 2023}

\normalsize

\author{ \href{https://orcid.org/0000-0002-9036-1031}{\includegraphics[scale=0.06]{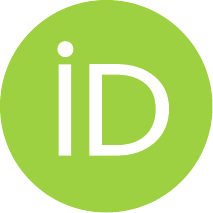}\hspace{1mm}Adrian ~Rojas-Campos} \\
	Institute of Cognitive Science\\
	Osnabrück University\\
	\texttt{rrojascampos@uos.de} \\
	\And
	\href{https://orcid.org/0000-0002-2958-0921}{\includegraphics[scale=0.06]{orcid.pdf}\hspace{1mm}Lukas~Stelz} \\
  Frankfurt Institute for Advanced Studies\\
	\texttt{stelz@fias.uni-frankfurt.de} \\
 	\And
	\href{https://orcid.org/0000-0003-0538-6670}{\includegraphics[scale=0.06]{orcid.pdf}\hspace{1mm}Pascal~Nieters} \\
	Institute of Cognitive Science\\
	Osnabrück University\\
	\texttt{pascal.nieters@uos.de} \\
}

\renewcommand{\shorttitle}{Learning COVID-19 Regional Transmission Using Universal Differential Equations}

\hypersetup{
pdftitle={A template for the arxiv style},
pdfsubject={q-bio.NC, q-bio.QM},
pdfauthor={David S.~Hippocampus, Elias D.~Striatum},
pdfkeywords={First keyword, Second keyword, More},
}

\begin{document}
\maketitle

\begin{abstract}
    Highly-interconnected societies difficult to model the spread of infectious diseases such as COVID-19. Single-region SIR models fail to account for incoming forces of infection and expanding them to a large number of interacting regions involves many assumptions that do not hold in the real world. We propose using Universal Differential Equations (UDEs) to capture the influence of neighboring regions and improve the model's predictions in a combined SIR+UDE model. UDEs are differential equations totally or partially defined by a deep neural network (DNN). We include an additive term to the SIR equations composed by a DNN that learns the incoming force of infection from the other regions. The learning is performed using automatic differentiation and gradient descent to approach the change in the target system caused by the state of the neighboring regions. We compared the proposed model using a simulated COVID-19 outbreak against a single-region SIR and a fully data-driven model composed only of a DNN. The proposed UDE+SIR model generates predictions that capture the outbreak dynamic more accurately, but a decay in performance is observed at the last stages of the outbreak. The single-area SIR and the fully data-driven approach do not capture the proper dynamics accurately. Once the predictions were obtained, we employed the SINDy algorithm to substitute the DNN with a regression, removing the black box element of the model with no considerable increase in the error levels.

\end{abstract}

% keywords can be removed

\newpage

\section{Introduction}

%% Epidemiology and Mathematical models %%
Epidemiology's central task is understanding infectious disease's occurrence, distribution, causes, and transmission. Mathematical models play a fundamental role in achieving this goal, describing and simulating disease dynamics, which helps to forecast and mitigate their transmission. As shown during the COVID-19 pandemic, the highly interconnected world we live in, where individuals often move between regions, exhibits transmission dynamics that challenge the traditional epidemiological models and make them difficult to model and predict. A new approach that combines Deep Neural Networks (DNNs) with differential equations could be a potential solution to this problem, providing flexibility by learning the incoming force of infection based on data and improving the predictive power of the model.

\begin{figure}[h]
    \centering
    \includegraphics[width=14cm]{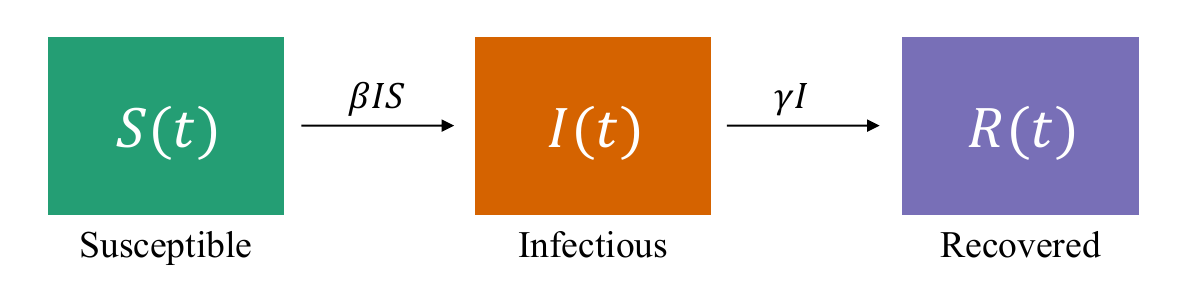}
    \caption{Diagram of the SIR model: each compartment is represented by a box and the movement of individuals between compartments as black arrows.}
    \label{fig:SIR_diagram}
\end{figure}

%% The SIR model %%
One of the main types of epidemiological models is compartmental models, where the population under study is divided into classes or compartments, and assumptions are made about the nature and time transfer rate from one class to others \parencite{Brauer_2019}. The Kermack-McKendrick model, also known as the SIR model, is the prototype compartmental model. Developed in 1927, it is considered one of the early triumphs of epidemiological modeling, generating predictions and insights about the behavior of natural epidemics. This model is based on simple assumptions on flow rates between compartments, as illustrated in Figure \ref{fig:SIR_diagram}. The population is divided into three classes which are functions of time $t$: susceptibles $S(t)$ is the number of individuals susceptible to the disease who have not yet been infected at time $t$, infectious $I(t)$ denotes the number of infected individuals who are infective and able to spread the disease by contact with susceptible; and recovered $R(t)$ is the number of individuals who have been infected and removed from the possibility of being infected \parencite{Kermack1927}.

The progression of individuals between compartments is controlled by two main parameters: $\beta$ is the transmission rate and controls the rate at which infectious people can spread the disease, with larger values implying faster spread and $\gamma$ is the recovery rate, which is the fraction of people that are removed from the infectious each time.

Additionally, the model needs to make assumptions about the disease's behavior: there is a constant rate of contact between compartments, an exponentially distributed recovery rate, no change in the total population, the sizes of the compartments are large enough that the mixing of members is homogeneous, and the transmission between compartments is deterministic in the population. Despite being based on assumptions that do not always hold, the model has proven helpful in modeling the spread of several diseases such as SARS and H1N1, and most recently, COVID-19 \parencite{Cooper2020, Brauer_2019}.

%% Derivations of the SIR model %%
Given its success, several derivations from the original model have been developed to assess additional information and different dynamics. For example, the SIS model allows individuals to be reinfected, the SIRD model contemplates the possibility of individuals dying from the epidemic, and the SEIR models are used for diseases that present a latency period between exposure and infection, among many others, resulting in a wide variety of applications \parencite{Hethcote1989}.

\subsection{Modeling COVID-19 dynamics}

%% Using SIR Models with covid19 %%
Compartmental models played a fundamental role in simulating, managing, and predicting outbreaks during the COVID-19 pandemic. Multiple publications applied SIR-based models to simulate the spread dynamics of entire regions or countries \parencite{Cooper2020, Perakis2023, Gounane2021, Marinov2022, Ianni2020}. Another prolific application area was to guide public interventions and management policies by simulating different possible scenarios based on the known dynamics \parencite{BarbarrosaFuhrmann2021, Barbarrosa2021Fleeing}.

%% Two problems: not considering neighboring regions and low-resolution %%
However, although we live in a highly interconnected world where individuals often move between regions, most applications of the SIR model focus on predicting single areas of interest, modeling at the level of entire cities or countries. The interactions between the spread dynamics of neighboring regions are often complex or unknown, challenging the simplification of single-area models. Additionally, modeling at the level of entire countries does not allow the capture of individual properties of each district caused by geographical or socioeconomic differences, which makes it impossible to generate file-grained predictions to guide localized interventions.

%% SIR model attempts to consider regional transmission %%
One approach to model regional transmission is the multi-group epidemic Lagrangian framework, which offers the possibility to model the risk of infections as a function of patch residence time and local environment risks. In this approach, the transmission is viewed as the product of contacts between individuals who move between patches with different risks of infection, and it depends on the time spent in each patch \parencite{Brauer_2019}. However, these models rely on oversimplified assumptions of mobility dynamics that are difficult to hold when the number of regions increases, which limits their application to real-world scenarios.

Only a handful of publications focus on modeling regional transmission. \textcite{Barbarrosa2021Fleeing} investigated the effect of the lockdown-generated fleeing from one region to another using an extended version of the SEIR model. \textcite{Goel2021} simulated the generation of a pandemic using the model of complex networks, representing separate regions in a grid with different transmission dynamics.

\subsection{Universal Differential Equations (UDE)}

%% Research goal: using UDE to complement SIR models %%
Given the difficulty presented by regional transmission and the unknown influence it can have over a target region, we propose to use and test a novel and promising Universal Differential Equations approach (UDEs) to tackle this problem. UDEs are differential equations fully or partially defined by a universal function approximator \parencite{rackauckas2021universal}, such as deep neural networks (DNNs). DNNs are capable of learning highly nonlinear functions based on data, by building a multi-level representation of the input and finding the parameters $\theta$ that result in a reasonable approximation \parencite{Goodfellow2016}.

\vspace{1cm}

A two dimensional UDE system with variables $x$ and $y$ may be described as:

\begin{equation}
\begin{aligned}
    \dot{x} &= f(x,y,t) + \text{DNN}_{\theta}(x,y) \\
    \dot{y} &= g(x,y,t) \\
\end{aligned}
\end{equation}

where the functions $f$ and $g$ describe the known dynamics for $x$ and $y$, respectively, and a new additive term is included in the dynamics of $x$ using a DNN$(x,y)$ with parameters $\theta$. The additive term DNN approximates a desired interaction based on the $x$ and $y$ dynamics. Using initial values for $x$, $y$, and $\theta$, the UDE can be numerically solved, and the trajectories for both variables can be estimated. The estimated trajectories based on the UDE can be compared against the actual trajectories at sample points $t$. Using automatic differentiation packages \parencite{bezanson2017, innes2019dont}, the parameters $\theta$ and the numerical solutions of the differential equations and the loss calculation can be differentiated. Based on this differentiation, $\theta$ can be optimized using gradient descent or a similar optimizer. This process results in a UDE with a parameterized DNN that optimally approximates the original time series of the system \parencite{rackauckas2021universal, Vortmeyer2021}

A couple of publications have previously implemented UDEs as part of SIR models. \textcite{Dandekar2020100145} used a UDE approach to augment the standard SIR model by calculating time-dependent quarantine effects on virus exposure using a DNN. On the other hand, \textcite{Kuwahara2023} implemented a UDE intending to predict COVID-19 second waves based on training an augmented SIR model only based on the first wave. To our knowledge, this is the first attempt to augment the model to capture regional transmission using a UDE approach.

\subsection{Augmenting COVID-19 modeling with UDEs}

The main idea of this study is to use a DNN to approximate the incoming force of infection from neighboring regions to a target region, improving the model's ability to capture the dynamics and its predictive capabilities. Concretely, we include a new term in the SIR model composed of a DNN that learns the influence of all neighboring regions over the target region and improves the prediction quality without making complex assumptions about the effect of such adjacent regions. Additionally, we employ the SINDy algorithm as a posthoc tool to provide an algebraic form to the learned function by the DNN, adding new terms to the model based on the training data.

%% UODEs and SIR %%
% There have been a couple of publications that previously combined SIR models and UDEs. \textcite{Dandekar2020100145} used a UDE approach to augment the standard SIR model by calculating time-dependent quarantine effects on virus exposure using a DNN. The DNN encodes the information about the quarantine strength and outputs an additive term to the infected compartment. The new term denotes the rate at which infected persons are effectively quarantined and isolated. This augmentation of the SIR model allowed the authors to better capture the dynamics and extract valuable information regarding the quarantine policies. Also, \textcite{Kuwahara2023} implements a UDE intending to predict COVID-19 second waves based on training an augmented SIR model only based on the first wave. In this case, the DNN included the mobility information inside each region. The implementation achieves moderate success and presents interesting ideas about improving the SIR model predictions, including external data to the model. To our knowledge, this is the first attempt to augment the model to capture regional transmission using a UDE approach.

%% Outlook %%
We start with a detailed description of the data and the models presented in the methods section. An evaluation and comparison between different models is offered in the results section. Finally, considerations and limitations of this study are presented in the discussion section.

\section{Methods}

We tested our proposed approach using synthetic data, which allows us to validate the test viability of the proposed ideas. For this, we simulate a COVID-19 outbreak in ten regions over a period of 500 days. The simulated outbreak provided the SIR dynamics for each of the ten regions based on each region's mobility and population density. We implemented, trained, and tested three model types for each region using the first half of the simulated outbreak (from day 1 to 250). The performance of each model was evaluated using the second half of the pandemic as test dataset (from day 251 to 500) and the complete pandemic (from day 1 to 500). The predictions are evaluated using the mean average error (MAE) and the Absolute Incidence Error (AIE). A detailed view of each of these steps is presented in the following.

\subsection{Simulated outbreak}

The synthetic data was generated using a SIR model on a graph, where each region is represented by one graph node. For each region $i$, the dynamics are governed by a set of SIR ordinary differential equations as in equation \ref{eqn:SIR}. The mean infectious period $\gamma^{-1}$ is equal in all regions. The force of infection $\lambda_i$ is dependent on intra- and inter-regional infections:

\begin{equation}
\begin{aligned}
\lambda_i = \frac{\beta_i I_i}{N_i} + \sum_{j\neq i} (M_{ij} + M_{ji})\frac{\beta_j I_j}{N_j}
\end{aligned}
\end{equation}

where $\beta_i$ and $N_i$ are the transmission rate and population size in region $i$, respectively. The transmission rate is equal in all regions: $\beta_i = \beta$. The mobility matrix $M_{ij}$ contains the weights of the edges of the bidirected mobility graph. Distance-based mobility was used to simulate the mobility between regions, which is dependent on both the distance $|x_i - x_j|$ between two regions and the population density of the destination region $\rho_j$:

\begin{equation}
\begin{aligned}
M_{ij} = f((1 + \frac{1}{2} \frac{\rho_j}{\sigma + \rho_j}) \cdot \frac{1}{|x_i - x_j|^2})
\end{aligned}
\end{equation}

where the function $f(x) = 1 - \frac{1}{(x + 1)^3}$ ensures that mobility does not diverge for regions that are very close together. The scale $\sigma$ controls the influence of population density.

\begin{figure}[h]
    \centering
    \includegraphics[width=4cm]{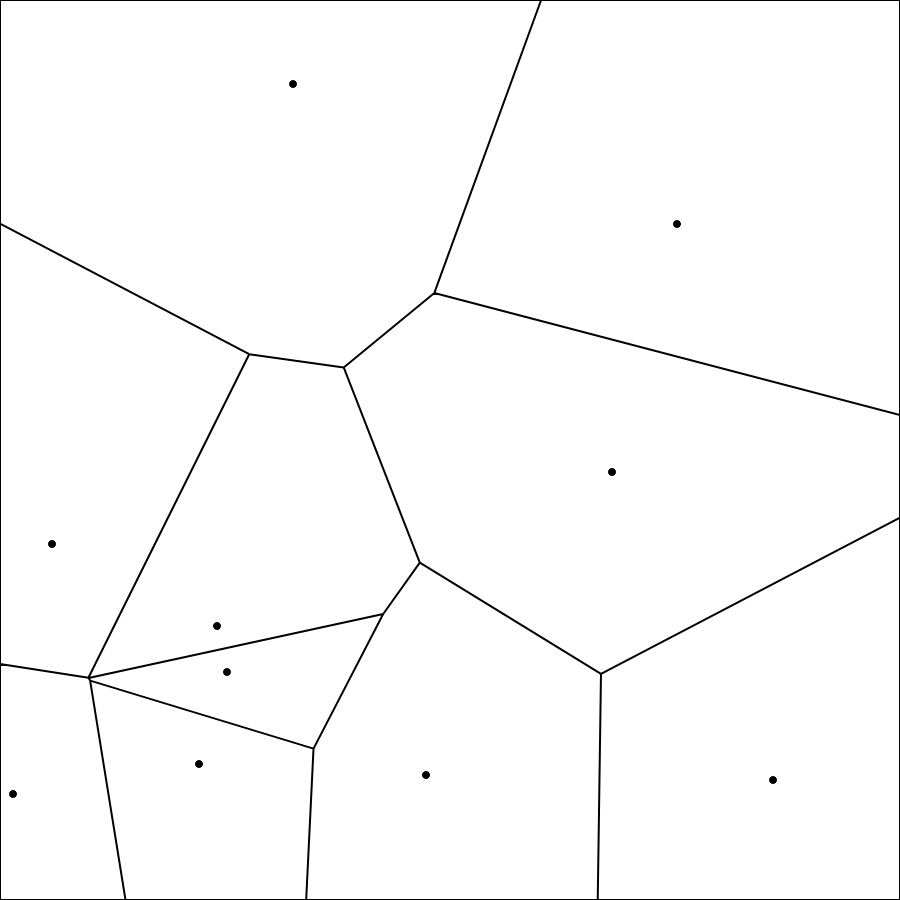}
    \caption{The outbreak was simulated using a Voronoi diagram of 10 regions with different properties. The mobility between regions was calculated based on the distance between regions and the population density. Twenty independent outbreaks were simulated based on twenty different initial conditions.}
    \label{fig:voronoi_cell_example}
\end{figure}

For the simulations, one random geography was generated for the ten regions by sampling $n$ points from a two-dimensional uniform distribution $\vec{x}_ i \sim U_2((0,4)^2)$. The total area is $A_{tot} = 4^2$ units. Each sampled point is used as the center of the corresponding Voronoi cell, similar to the example in Figure \ref{fig:voronoi_cell_example}. The population size of each region is sampled from a log-uniform distribution on the interval $[10^3,10^4]$: $\ln(N_i) \sim U(\ln(10^3), \ln(10^4))$. Finally, the population density is calculated using the actual area of the Voronoi cells.

A total of 40 different random initial conditions are prepared based on two possible scenarios. In the first 20, no initially recovered individuals were considered. Ten initially infected individuals are distributed to the ten regions. The region's population size weights the probability of having an infected individual. In the second scenario, also with 20 random initial conditions, a quarter of the total population was assumed to have recovered. These individuals are distributed to the different regions the same way as the initially infected.

\begin{table}[h]
\centering
\begin{tabular}{l|c}
Simulated scenario & Dataset's shapes \\
\hline
No initially recovered &  20 initializations $\times$ 10 regions $\times$ 3 compartments $\times$ 500 days \\
\hline
25\% initially recovered & 20 initializations $\times$ 10 regions $\times$ 3 compartments $\times$ 500 days  \\
\end{tabular}
\\
\caption{\label{tab:sets}Shape of the datasets for each simulated scenario.}
\end{table}

\begin{figure}[h]
    \centering
    \includegraphics[width=12cm]{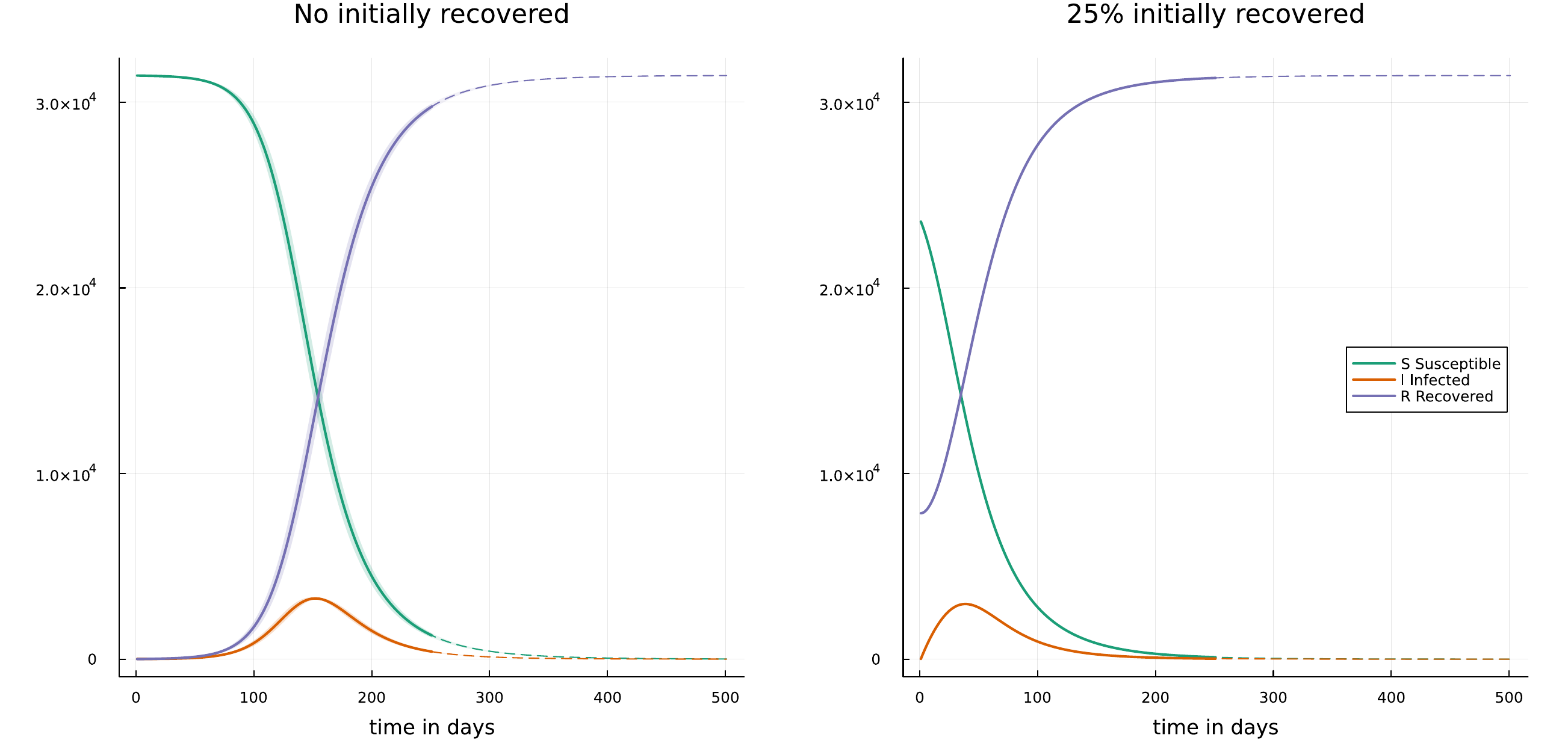}
    \caption{Accumulated cases over all regions in both simulated scenarios. Variations between different initializations are shown as a colored area.}
    \label{fig:train_test_sets_example}
\end{figure}

For each initial random condition, the SIR infection dynamics are simulated once for every mobility model over 500 days. The optimization scheme reports all compartments with a time step $dt = 1$ day. The simulated outbreak resulted in the datasets shown in Table \ref{tab:sets}. The parameters of the models were fitted using only the first half of the simulated outbreak (from day 1 to 250), as illustrated in Figure \ref{fig:train_test_sets_example}.

\subsection{Models}

For each of the ten regions of the simulated outbreak, we trained three models using the first 250 days. The interest region is designated as \textit{target} and the nine remaining as \textit{neighbors}. No model had access to the real parameters $\beta$ and $\gamma$ used to generate the simulations, but only the state of the SIR compartments at time \textit{t} is designated as $u_{\text{target}}(t)$ and the state of the nine neighboring regions as $u_{\text{neighbors}}(t)$. The complete process was repeated over each random initialization.

\subsubsection{SIR model: Parameter Fit with Gradient Descent}

As a baseline, we fit a single SIR model's infection rate $\beta$ and recovery rate $\gamma$, excluding the state from neighboring regions. Due to this omission, this model is not expected to generate the correct predictions for each location. Both parameters were initialized randomly. The mean squared error was used as a loss function, and the Adam optimizer with a learning rate of 0.01 was utilized. The trained model can be presented as:

\begin{equation}
\begin{aligned}
\dot{S}_{\text{target}} &= -\beta S I  \\
\dot{I}_{\text{target}} &= \beta S I - \gamma I\\
\dot{R}_{\text{target}} &= \gamma I \\
\end{aligned}
\label{eqn:SIR}
\end{equation}

\subsubsection{Full UDE model}

Additionally, we trained a universal differential equations model composed exclusively of a DNN. The DNN was trained to predict the changes in the target region, tacking as input information from the target and the neighboring regions. The DNN comprises an input layer, 16 hidden units with \textit{tanh} as an activation function, and an output layer with three output units, one for each model compartment of the target region. The weights and biases $\mu$ were initialized randomly and fitted during training. Given that different target locations have different sizes and the remaining neighboring regions change, the variables of the target location and the state of the adjacent areas were normalized between 0 and 1. The mean squared error was used as a loss function, and the Adam optimizer with a learning rate of 0.01 was utilized. This model can be summarized as follows, with $u$ as the state of the SIR system at a specific time $t$:

%% Description of the ANN %%

\begin{equation}
\begin{aligned}
\dot{S}_{\text{target}} &= \text{DNN}_{\mu}(u_{\text{target}};u_{\text{neighbors}})[1] \\
\dot{I}_{\text{target}} &= \text{DNN}_{\mu}(u_{\text{target}};u_{\text{neighbors}})[2] \\
\dot{R}_{\text{target}} &= \text{DNN}_{\mu}(u_{\text{target}};u_{\text{neighbors}})[3] \\
\end{aligned}
\end{equation}

\subsubsection{SIR + UDE model}

Our model of interest combines the SIR model with an extra additive term estimated by a DNN. This additive term captures the influence of the state of other regions and adds it to the infected compartment while subtracting it from the susceptibles.  In this case, the DNN is composed of an input layer that takes the states of the neighboring regions normalized between 0 and 1, a hidden layer with 16 units with \textit{tanh} activation function, and a single output layer with \textit{softplus} activation function with a positive output that ensures stability of the system during training. The goal of this DNN is to approximate the influence of neighboring regions to the target system, assuming that the influence of the neighboring regions is positive and additive and positive to the infected compartment and negative to the susceptibles. During the optimization process, the parameters of the SIR model ($\beta$, $\gamma$), as well as the parameters of the DNN $\theta$, are learned. The mean squared error was used as a loss function. The training was performed in two steps: first using Adam optimizer with a learning rate of 0.001, and later with BFGS with a learning rate 0.01. This model is described as follows:

\begin{equation}
\begin{aligned}
\dot{S}_{\text{target}} &= -\beta S I - \text{DNN}_{\theta}(u_{\text{neighbors}})\\
\dot{I}_{\text{target}} &= \beta SI + \text{DNN}_{\theta}(u_{\text{neighbors}})- \gamma I  \\
\dot{R}_{\text{target}} &= \gamma I \\
\end{aligned}
\end{equation}

\subsection{Using SINDy to Remove the Black Box}

Using a DNN as a part of the UDE model can improve the model's predictions but does not provide any information about how the neighboring regions affect the target region. Previous publications used the Sparse Identification of Nonlinear Dynamics (SINDy) algorithm in the context of UDEs \parencite{Brunton2016} to find the right algebraic expressions behind dynamical systems data firstly approximated by a DNN, successfully removing the black box from the equation \parencite{Vortmeyer2021, rackauckas2021universal}. Therefore, we use the SINDy algorithm to perform a post hoc analysis of the function learned by the DNN, intending to discover an algebraic form for such a function and substitute the DNN element of the SIR+UDE model.

SINDy helps to determine the governing equations from dynamical systems data by leveraging that most physical systems have only a few relevant terms that define the dynamics, making the equations sparse in a high dimensional nonlinear space. This algorithm formulates finding the algebraic differential equation form as a linear regression problem: a matrix of non-linear functions $\Theta$ applied to the state vector $x$ of the system multiplied by a matrix of sparse coefficient vectors $\Xi$ recovers the standard form of many non-linear dynamics in a way $\dot{x} = f(x) = \Theta (x) \Xi$. If measurements for both $x$ and $\dot{x}$ are available at several sample points in time, we can write $\dot{X} = \Theta (X) \Xi + \eta Z $ where X and $\dot{X}$ have rows for each sample point, $\Theta(X)$ is a design matrix of non-linear functions applied to the data and $\eta Z$ is independent and identically distributed Gaussian noise with magnitude $\eta$. This is a standard linear regression problem in multiple variables,  where finding the sparse coefficient vectors $\Xi = [\xi_{1}, ... ,\xi_{n}]$ for a system with $n$ state variables by applying the LASSO algorithm can find those non-linear functions $x$ in the design matrix that best explains the data \parencite{Vortmeyer2021, Brunton2016}.

In our implementation, we select the SIR+UDE model for both initial scenarios with the lowest error over the train and test sets and apply the SINDy algorithm to the input and output of the DNN at each region for the complete evolution of the pandemic. We then use the regression result to reproduce the outbreak and measure the error levels to compare them with the DNN. The SIR+SINDy model can be defined with $\omega$ as the learned weights and $b$ the bias:

\begin{equation}
\begin{aligned}
\dot{S}_{\text{target}} &= -\beta S I - (u_{\text{neighbors}} * \omega + b) \\
\dot{I}_{\text{target}} &= \beta SI + (u_{\text{neighbors}} * \omega + b) - \gamma I  \\
\dot{R}_{\text{target}} &= \gamma I \\
\end{aligned}
\end{equation}

\subsection{Metrics}

Two different metrics were used to evaluate the quality of the predictions on the unobserved timesteps. On one side, a machine learning-oriented metric, such as the mean absolute error (MAE) and a domain-specific evaluation as absolute incidence's error (AIE).

\subsubsection{Mean Absolute Error (MAE)}

The mean absolute error (MAE) is one of the standard metrics in the deep learning field, and it is defined as the average absolute difference between the estimated $\hat{Y}$ and the actual values $Y$ (including the predictions for all three compartments) using the formula:

\begin{equation}
\begin{aligned}
\text{MAE} &= \frac{1}{n} \sum_{i=1}^{n} \lvert Y_{i} - \hat{Y}_{i} \lvert  \\
\end{aligned}
\end{equation}

\subsubsection{Absolute Incidence Error (AIE)}

Additionally, we evaluate the predictions of the models with a metric based on the daily incidence $\lambda$ in percentages:

\begin{equation}
\begin{aligned}
\lambda  &= \frac{\text{New infected in 1 day}}{\text{Total population}} * 100  \\
\end{aligned}
\end{equation}

We define the absolute incidence's error (AIE) as the sum of the absolute difference between the estimated $\hat{\lambda}$ and the observed incidence $\lambda$ over the complete array, calculated such as:

\begin{equation}
\begin{aligned}
\text{AIE} &= \sum_{i=1}^{n} \lvert \lambda_{i} - \hat{\lambda}_{i} \lvert  \\
\end{aligned}
\end{equation}

\subsection{Implementation} %%% TO DO: Mention SINDY %%%

The models were trained on the JUWELS Supercomputer at Jülich Supercomputing Centre (JSC), using accelerated compute nodes. Each node contained 2 Intel Xeon Platinum 8168 CPU with 20 cores each of 2.7 GHz and 192 GB DDR4 of RAM at 2666 MHz. The implementation was based on the Julia Programming Language V1.7.1 \parencite{bezanson2017}. The SciML environment was used to implement and train the differential equations models \parencite{rackauckas2021universal}, specifically the libraries: SciMLBase v1.26.1, ModelingToolkit v8.4.0, DataDrivenDiffEq v0.7.0 and DiffEqFlux v1.45.0. Array storage was performed with JLD v0.13.1, Interpolations v0.13.5 manage the interpolation of neighboring status to undefined timepoints, and Plots v1.25.8 manage the plotting of results. MPI.jl distributed the jobs in the HPC system \parencite{Byrne2021}.  The complete implementation can be found in the repository \url{https://github.com/adrocampos/UDEs_in_SIR_regional_transmision}.

\section{Results}

The evaluation of the three different models in both scenarios is presented in this section. We report the median values over the 20 random initializations in each scenario (no initially recovered and 25\% initially recovered). The models' predictions were evaluated in two distinct ways: measuring the difference between the predictions for the second half of the pandemic (test set) with the goal of measuring each model's ability to accurately predict the evolution of the pandemic, and the difference in reproducing the complete progression of the outbreak (train and test sets) to evaluate the ability of the models to represent the outbreak correctly.

\subsection{Predicting the evolution of the pandemic: test set}

\begin{figure}[b!]
    \centering
    \includegraphics[width=16cm]{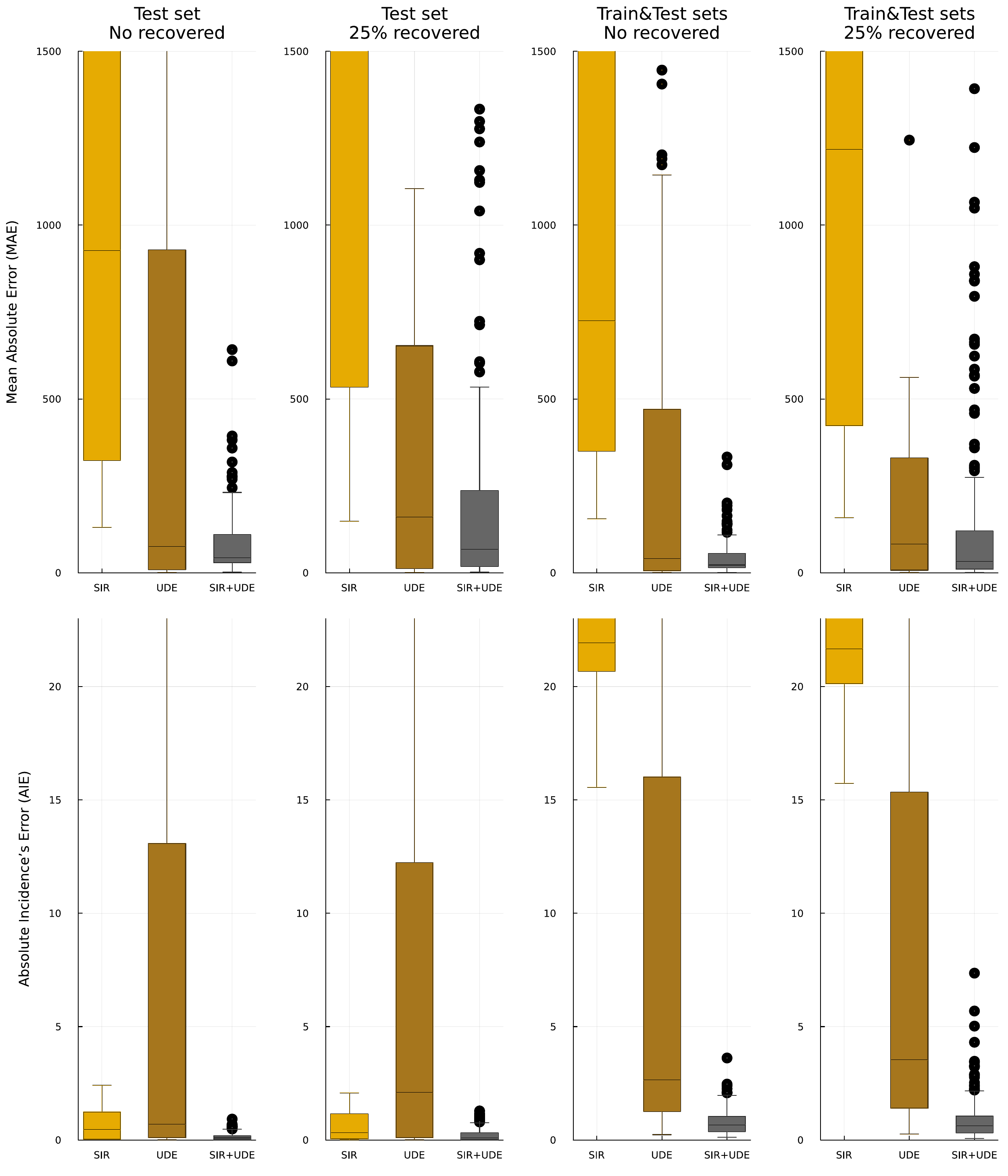}
    \caption{Overall performance comparison between models using both metrics at each simulated scenario. Each bar plot illustrates the performance of overall initializations and regions. The median is presented as a horizontal line.}
    \label{fig:model_comparison}
\end{figure}

%%% MAE %%%
We first evaluate the models' predictions to the second half of the outbreak, intending to asses the generalizability of the models in unobserved data. As illustrated in Figure \ref{fig:model_comparison}, the SIR+UDE model obtained the lowest overall median MAE, followed by the UDE and the SIR. Analyzing the performance at each specific region in Table \ref{tab:results_test_set}, we see that the SIR+UDE model has the lowest median MAE on most regions, but the UDE overperforms it in a few regions. We can also observe that the UDE's median MAE levels strongly vary between regions. The SIR model shows the highest MAE median in most regions.

%%% AIE %%%
An interesting scenario is found in observing at the AIE measurements. The UDE+SIR model shows the lowest median AIE measurements overall regions, seconded by the SIR model and the UDE with the highest error level. Analyzing the values for each specific region, we found that the SIR+UDE has the lowest error values in most regions, while the UDE shows the highest median AIE in most regions, showing some large differences between regions. The SIR's median AIE measurements are low and close to the UDE+SIR performance in most regions.

\begin{table}[h!]
    \centering
    \begin{tabular}{cc|ccc|ccc}
    \multicolumn{2}{c}{} & \multicolumn{3}{c}{\textbf{No initially recovered}} & \multicolumn{3}{c}{\textbf{25\% initially recovered}}  \\
    & Region & SIR & UDE & UDE+SIR & SIR & UDE & UDE+SIR  \\
    \hline
    \parbox[t]{2mm}{\multirow{11}{*}{\rotatebox[origin=c]{90}{\textbf{Mean Absolute Error (MAE)}}}}
     & \textbf{All} & \textbf{927.5} & \textbf{76.3} & \textbf{42.2} & \textbf{1588.4} & \textbf{161.3} & \textbf{66.6} \\
     & 1 & 733.0 & 1025.3 & 122.7 & 3944.5 & 507.8 & 59.6 \\
     & 2 & 2028.2 & 8.0 & 27.0 & 153.3 & 13.2 & 7.9 \\
     & 3 & 2911.7 & 414.5 & 161.4 & 2479.9 & 5762.8 & 1654.8 \\
     & 4 & 2199.1 & 4.0 & 32.9 & 1209.5 & 100.5 & 248.4 \\
     & 5 & 780.9 & 1236.5 & 89.4 & 3777.8 & 583.2 & 110.9 \\
     & 6 & 279.3 & 11.7 & 38.1 & 2829.3 & 6.0 & 17.1 \\
     & 7 & 3202.9 & 356.2 & 76.5 & 2890.9 & 313.6 & 114.0 \\
     & 8 & 2391.9 & 7.0 & 5.0 & 2133.3 & 3.2 & 12.8 \\
     & 9 & 961.4 & 1903.3 & 168.5 & 1477.2 & 3675.2 & 450.9 \\
     & 10 & 321.1 & 19.3 & 32.7 & 540.5 & 20.1 & 57.4 \\
    \hline
    \parbox[t]{2mm}{\multirow{11}{*}{\rotatebox[origin=c]{90}{\textbf{Absolute Incidence’s Error (AIE)}}}}
     & \textbf{All} & \textbf{0.454} & \textbf{0.684} & \textbf{0.092} & \textbf{0.312} & \textbf{2.089} & \textbf{0.105} \\
     & 1 & 1.063 & 13.737 & 0.189 & 0.774 & 11.024 & 0.12 \\
     & 2 & 0.027 & 0.099 & 0.007 & 0.008 & 0.268 & 0.003 \\
     & 3 & 1.865 & 7.347 & 0.59 & 1.058 & 23.358 & 0.876 \\
     & 4 & 0.197 & 0.044 & 0.078 & 0.197 & 1.0 & 0.101 \\
     & 5 & 1.269 & 17.057 & 0.149 & 1.323 & 10.466 & 0.268 \\
     & 6 & 0.031 & 0.073 & 0.009 & 0.048 & 0.045 & 0.012 \\
     & 7 & 0.976 & 8.965 & 0.176 & 1.231 & 6.054 & 0.318 \\
     & 8 & 0.014 & 0.125 & 0.003 & 0.056 & 0.039 & 0.014 \\
     & 9 & 1.054 & 25.697 & 0.162 & 0.933 & 27.009 & 0.35 \\
     & 10 & 0.014 & 0.132 & 0.004 & 0.037 & 0.097 & 0.011 \\
    \hline
    \\
    \end{tabular}
    \caption{\label{tab:results_test_set} Medians of error measurements over random initializations in test set (from day 1 to 250)}
\end{table}

\subsection{Learning the dynamics of the complete pandemic: train and test set}

%%% MAE %%%
In the second step, we evaluate the predictions of the models for the complete outbreak, intending to measure the model's ability to abstract the phenomena' right dynamics. Figure \ref{fig:model_comparison} shows that the median MAE values for the complete pandemic are similar to the previously observed values for the evolution of the pandemic: the SIR+UDE obtained the lowest median MAE values, followed by the UDE and SIR at last. Observing the values at each region at Table \ref{tab:results_all_sets}, we see that the SIR+UDE model also has the lowest median MAE at most regions, but the UDE obtained a lower median error in a minority of regions. The same strong variability between regions in UDE error levels is seen. The SIR model shows the highest median MAE in almost all regions.

%%% AIE %%%
Analyzing the AIE median values, we observe that the advantage of the SIR model over the UDE disappears. The SIR+UDE shows the lowest median AIE values, followed by the UDE and the SIR models. The same pattern is found when observing the specific region performance, where the SIR+UDE obtained the lowest AIE values in all stations and the SIR model the highest in most of them.

\begin{table}[h!]
    \centering
    \begin{tabular}{cc|ccc|ccc}
    \multicolumn{2}{c}{} & \multicolumn{3}{c}{\textbf{No initially recovered}} & \multicolumn{3}{c}{\textbf{25\% initially recovered}}  \\
    & Region & SIR & UDE & UDE+SIR & SIR & UDE & UDE+SIR  \\
    \hline
    \parbox[t]{2mm}{\multirow{11}{*}{\rotatebox[origin=c]{90}{\textbf{Mean Absolute Error (MAE)}}}}
    & \textbf{All} & \textbf{724.0} & \textbf{40.0} & \textbf{22.3} & \textbf{1219.1} & \textbf{83.4} & \textbf{33.7} \\
    & 1 & 569.7 & 518.2 & 64.1 & 2787.4 & 257.9 & 30.4 \\
    & 2 & 1864.2 & 4.9 & 14.3 & 187.0 & 7.5 & 4.4 \\
    & 3 & 1921.0 & 208.3 & 81.2 & 1645.2 & 2910.1 & 839.7 \\
    & 4 & 1907.7 & 2.4 & 16.8 & 1155.1 & 54.0 & 126.8 \\
    & 5 & 613.6 & 627.1 & 46.3 & 2575.5 & 294.3 & 56.6 \\
    & 6 & 320.6 & 6.7 & 20.2 & 2540.8 & 3.5 & 8.9 \\
    & 7 & 2241.9 & 180.3 & 38.8 & 1993.0 & 158.0 & 57.7 \\
    & 8 & 2183.7 & 4.0 & 2.7 & 1910.7 & 2.0 & 6.7 \\
    & 9 & 750.2 & 967.4 & 85.7 & 1087.8 & 1866.9 & 228.1 \\
    & 10 & 389.4 & 11.3 & 17.9 & 614.7 & 12.5 & 30.0 \\
    \hline
    \parbox[t]{2mm}{\multirow{11}{*}{\rotatebox[origin=c]{90}{\textbf{Absolute Incidence’s Error (AIE)}}}}
    & \textbf{All} & \textbf{21.914} & \textbf{2.639} & \textbf{0.677} & \textbf{21.676} & \textbf{3.531} & \textbf{0.636} \\
    & 1 & 21.74 & 17.031 & 0.929 & 23.2 & 13.772 & 0.703 \\
    & 2 & 30.092 & 1.383 & 0.403 & 44.203 & 1.417 & 0.285 \\
    & 3 & 19.033 & 8.049 & 1.098 & 16.945 & 27.17 & 2.208 \\
    & 4 & 16.072 & 0.504 & 0.323 & 34.68 & 2.515 & 0.515 \\
    & 5 & 21.313 & 20.397 & 0.839 & 21.077 & 12.204 & 0.913 \\
    & 6 & 39.572 & 1.326 & 0.433 & 19.609 & 0.809 & 0.239 \\
    & 7 & 22.566 & 10.98 & 0.843 & 21.206 & 7.233 & 0.953 \\
    & 8 & 21.932 & 1.118 & 0.289 & 19.494 & 0.776 & 0.292 \\
    & 9 & 22.621 & 31.983 & 1.331 & 21.118 & 33.371 & 1.257 \\
    & 10 & 44.066 & 1.755 & 0.663 & 41.703 & 1.706 & 0.36 \\
    \hline
    \\
    \end{tabular}
    \caption{\label{tab:results_all_sets} Medians of error measurements over 20 random initializations for the train \& test sets (from day 1 to 500)}
\end{table}

\subsection{Predicting each compartment}

\begin{figure}[h!]
    \centering
    \includegraphics[width=16cm]{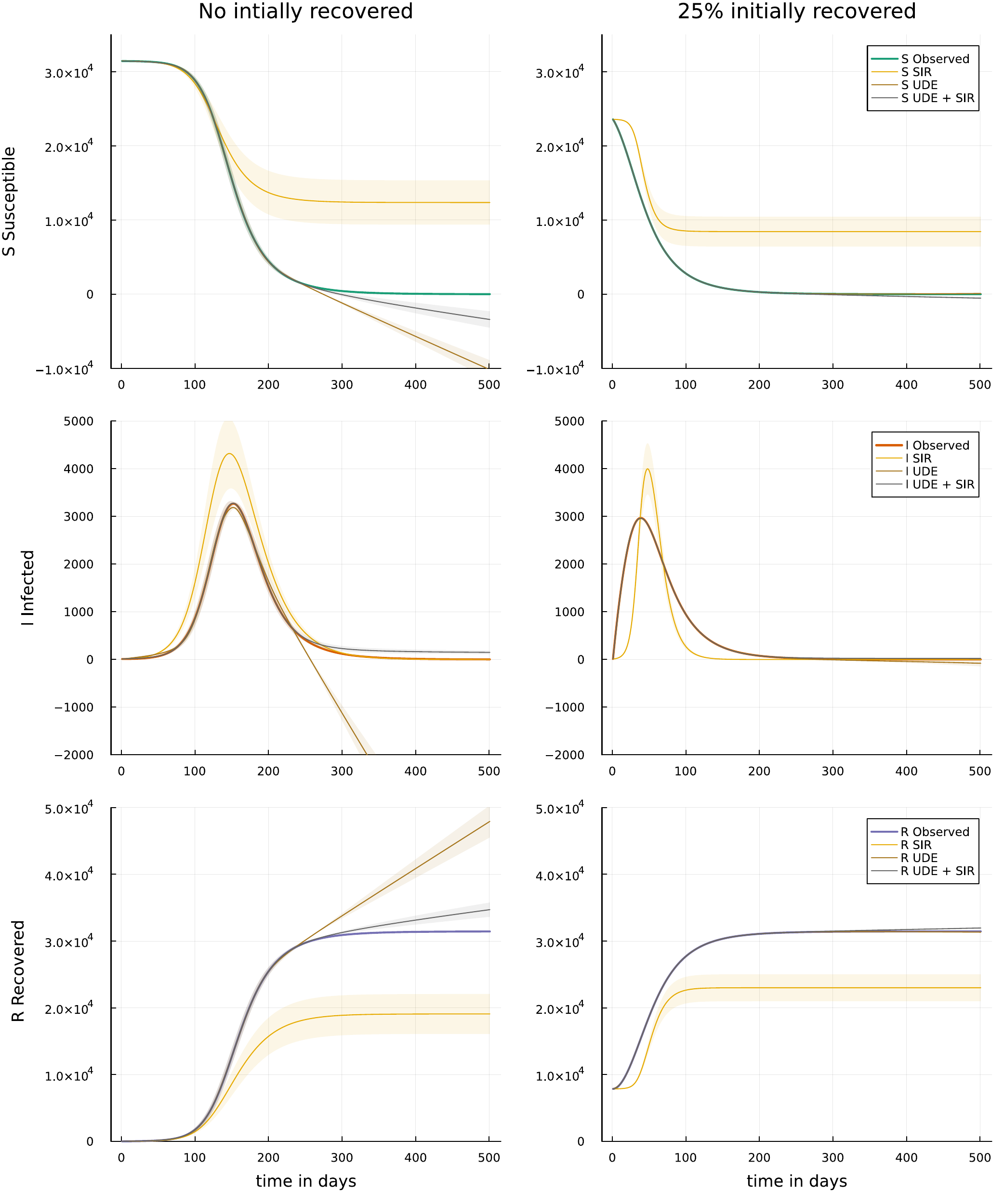}
    \caption{Cumulative predictions over all regions at each compartment. The predictions for the different initializations are illustrated as colored areas and the median of the initializations as a solid line.}
    \label{fig:predictions}
\end{figure}

The cumulative predictions over all areas are plotted in Figure \ref{fig:predictions}. In these plots, we can see that the SIR+UDE model best captures the pandemic's dynamics and makes the closest predictions to the simulated values in all compartments. However, an important decay in performance is observed at the end of the outbreak, where the predictions progressively deviate from the simulated data with time.

The predictions of the UDE models show a very good fitting to the train set, but almost no skill to estimate the evolution of the pandemic in the ``No initially recovered'' scenario. This is not the case when 25\% of the population is recovered, where the UDE model achieves good predictions. However, in this scenario, after 250 days the outbreak is mostly over, which enable the DNN to learn all meaningful relationships in the data, and no extrapolation is needed.

The SIR model, on the other hand, fails to estimate correctly all compartments. An important difference can be appreciated in the susceptible and recovered compartments. In the case of the infected compartment, we can see a significant overestimation of the peak, which causes a sudden drop of the infected for the rest of the pandemic, which reflects in good AIE values for the evolution of the pandemic but has an overall bad fit.

\subsection{SINDy results}

\begin{figure}[h!]
    \centering
    \includegraphics[width=12cm]{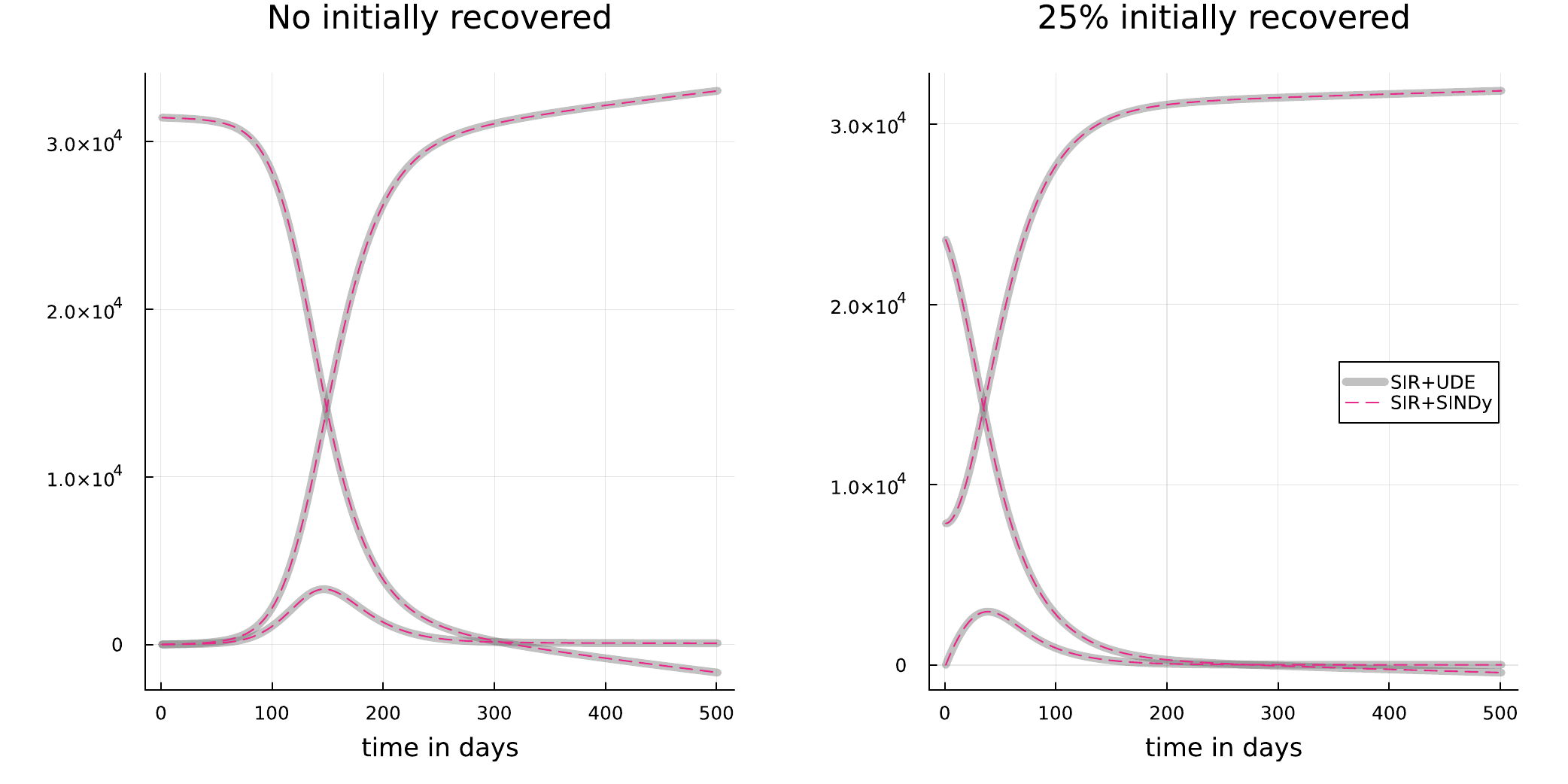}
    \caption{Comparison between cummulative predictions of the best UDE+SIR model and the UDE+SINDy predictions}
    \label{fig:sindy_reconstruction}
\end{figure}

Evaluating the difference between the error measurements of the SIR+UDE and the SIR+SINDy model presented in Table \ref{tab:results_sindy}, we appreciate a marginal difference between the performance of the two. This marginal difference and the comparison of the resulting predictions in Figure \ref{fig:sindy_reconstruction} suggests that a regression equation can substitute the DNN without compromising the quality of the predictions. Analyzing the regression coefficients makes it possible to interpret the influence of neighboring regions over each target region.

\begin{table}[h!]
    \centering
    \begin{tabular}{ccc|cc}
    \multicolumn{1}{c}{} & \multicolumn{2}{c}{\textbf{No initially infected}} & \multicolumn{2}{c}{\textbf{25\% initially infected}} \\
    Metric & SIR+UDE & SINDy & SIR+UDE & SINDy  \\
    \hline
    \multicolumn{1}{c}{} & \multicolumn{4}{c}{Results in test set}\\
    \textbf{MAE} & 1506.6 & 1497.8 & 412.3 & 412.6 \\
    \textbf{AIE} & 2.365 & 2.469 & 0.213 & 0.214 \\
    \multicolumn{1}{c}{} & \multicolumn{4}{c}{Results in train \& test sets}\\
    \textbf{MAE} & 767.6 & 818.3 & 212.7 & 212.9 \\
    \textbf{AIE} & 10.886 & 11.509 & 2.955 & 2.955 \\
    \hline
    \\
    \end{tabular}
    \caption{\label{tab:results_sindy} Error levels of the best SIR+UDE model and the SIR+SINDy predictions.}
\end{table}

\section{Discussion}

The main goal of this project was to offer an approach to how DNNs can be integrated into a SIR model to include regional transmission without needing a large number of assumptions. We test using UDEs to augment the SIR model to include the influence of neighboring regions and improve the model's predictions. To test the benefits of this approach, we compare its predictions against a single area SIR model and a complete data-driven UDE model, measuring the difference to the simulated values.

First, our results show that the SIR+UDE model was able to learn the influence of the neighboring regions over the target region through an additive term in the S and I compartments. The SIR+UDE model obtained the lowest median error levels of all models and generated the best guesses about the past and future behavior of the outbreak. However, the quality of the predictions decays in the long-term predictions for some compartments. The predicted median deviates from the real values as the predictions move away from the initial 250 days used to adjust the parameters.

%%% Hypothesis of why this happen? %% Idealed discussion phenomena?  %%% How to solve in the future?
We believe that the good performance of this model is due to the combination of knowledge from the physical phenomena provided by the SIR model and the data-driven learned term. The learned term captures the influence of neighboring regions in the training set and achieves good predictions for the following days of the pandemic. However, given that the training set was very different to the data points at the end of the outbreak, the learned information does not extrapolate correctly to the last unobserved period.

On the other hand, a purely data-driven approach composed only of a DNN (UDE models) could not extrapolate the information learned in the training set to predict the future states of the pandemic accurately. Given that this model is composed exclusively of a DNN with no physical knowledge about the evolution of the outbreak, it is not expected to generate quality predictions for the second half of the pandemic, considering the lack of similar training examples and the poor performance of multilayer perceptrons in sequential data \parencite{Goodfellow2016}.

As expected, using a single SIR model was insufficient to predict the outbreak correctly. A single SIR model for each region lacked enough information to estimate the correct values of its parameters. During parameter estimation, the model adjusted its parameters to approximate the course of the pandemic, which resulted in an overestimation of the magnitude of the pandemic and a bad general fit at each compartment. These results corroborate the limitations of a single modeling level in a high-interconnected environment.

Lastly, using the SINDy algorithm after performing the predictions allowed us to determine the influence of neighboring regions on the target and add coefficients to each neighboring region, elaborating a post hoc detailed model for each region and the influence from their neighboring regions.

%% How are these findings related to the antecedents? %%
In general, the results support the implementation of UDEs to augment the SIR model, especially when the new term is unknown and can not be easily calculated. Our results also show the advantages of opting for a mixed model that integrates physical knowledge with data-driven elements instead of the complete data-driven approach. We showed that using the SIR+UDE model allows us to estimate a model for each region that improves the quality of its predictions by including the influence of the neighboring regions without needing the values for mobility between regions. Using the SINDy algorithm as a post hoc tool, we were able to eliminate the black box element and provide algebraic terms for the influence of neighboring regions. This work offers a novel and experimental method to approach the problem of modeling the regional transmission of COVID-19 in many regions, and it forms part of a growing body of publications using UDEs to complement dynamical systems in contexts where large amounts of data are available \parencite{Nogueira2022, Kuwahara2023, Dandekar2020100145}.

%% Limitations %%
This work's most significant limitation is the use of synthetic data, which limits its direct application to real-world COVID-19 records. Real-world COVID-19 records can present noise and artifacts (errors of measurements, unreported cases, weekend effects, etc.) that require careful implementation of the ideas presented here, but future research should replicate the results in real-world scenarios and test the proposed approach using real data.

Another limitation is that the model assumes that the effect of the neighboring regions over a target system is additive. The main assumption is that the incoming force of infection is subtractive to the S compartment and additive to the I compartment, and the output of the DNN was always positive. These decisions were made to ensure stability during the training of the algorithms, and although they are theoretically coherent, there are nevertheless assumptions about how the system behaves.

Additionally, the decision to select the outbreak's first half as the train set and the second half as the test set is inconsistent with usual epidemiological applications. The main idea behind this decision was to test the extrapolation capabilities of each model over an extended period and to get information about the ability to predict the complete evolution of the outbreak rather than to generate useful short-term predictions that guide interventions. Future implementations of our findings using real-world COVID-19 records should also implement shorter prediction ranges that imitate the real-world applications of these models.

We see great potential in using UDEs to complement SIR models in real-world scenarios for short-term predictions. Implementing UDEs to integrate regional transmission (or other complex and unknown interactions) could help generate high-quality predictions in the very short term (one or two weeks) without needing to explicitly know and describe the complex mobility patterns that neighboring regions can exhibit. In general, integrating DNNs to solve specific limitations of the SIR models could provide flexibility to the modeling process that allows for fast and targeted interventions, and applying tools like SINDy for a post hoc analysis can help to analyze the learned term to contribute to a better understanding of the phenomena and provide insights for the theory. The combined use of UDEs and SINDy is a promising approach to tackling the complexity of real-world scenarios in a data-driven way, while SINDy sheds light to the phenomena and provide feedback for knowledge discovery.

\newpage
\printbibliography

@book{Brauer_2019,  
    title={Mathematical Models in Epidemiology}, 
    DOI={https://doi.org/10.1007/978-1-4939-9828-9}, 
    publisher={Springer New York, NY}, 
    author={Fred Brauer and Carlos Castillo-Chavez and Zhilan Feng}, 
    year={2019}
}

@misc{rackauckas2021universal,
      title={Universal Differential Equations for Scientific Machine Learning}, 
      author={Christopher Rackauckas and Yingbo Ma and Julius Martensen and Collin Warner and Kirill Zubov and Rohit Supekar and Dominic Skinner and Ali Ramadhan and Alan Edelman},
      year={2021},
      eprint={2001.04385},
      archivePrefix={arXiv},
      primaryClass={cs.LG}
}

@article{Brunton2016,
	title = {Discovering governing equations from data by sparse identification of nonlinear dynamical systems},
	pages = {3932--3937},
	doi = {https://doi.org/10.1073/pnas.1517384113},
	author = {Steven Brunton and Joshua Proctor and J. Nathan Kutz},
	number = {113},
	journal = {Proceedings of the National Academy of Sciences (PNAS)},
	volume = {15},
	year = {2016}
}

@article{BarbarrosaFuhrmann2021,
  author  = {Maria Vittoria Barbarrosa and Jan Fuhrmann}, 
  title   = {Germany’s next shutdown—Possible scenarios and outcomes},
  journal = {Influenza and other respiratory viruses},
  year    = 2021,
  number  = 15,
  pages   = {326-330},
  doi = {https://doi.org/10.1111/irv.12827}
}

@article{Barbarrosa2021Fleeing,
    doi = {https://doi.org/10.1038/s41598-021-88204-9},
    author = {Maria Vittoria Barbarossa and Norbert Bogya and Attila Dénes and Gergely Röst and Hridya Vinod Varma and Zsolt Vizi},
    journal = {Nature Scientific Reports},
    title = {Fleeing lockdown and its impact on the size of epidemic outbreaks in the source and target regions – a COVID-19 lesson},
    year = {2021},
    volume = {11},
    url = {https://www.nature.com/articles/s41598-021-88204-9},
    number = {9233},
}

@Inbook{Hethcote1989,
    author="Hethcote, Herbert W.",
    editor="Levin, Simon A.
    and Hallam, Thomas G.
    and Gross, Louis J.",
    title="Three Basic Epidemiological Models",
    bookTitle="Applied Mathematical Ecology",
    year="1989",
    publisher="Springer Berlin Heidelberg",
    address="Berlin, Heidelberg",
    pages="119--144",
    isbn="978-3-642-61317-3",
    doi="10.1007/978-3-642-61317-3_5",
    url="https://doi.org/10.1007/978-3-642-61317-3_5"
}

@article{Kermack1927,
    doi = {https://doi.org/10.1098/rspa.1927.0118},
    author = {William Ogilvy Kermack and A. G. McKendrick},
    journal = {Proc. Royal Soc. London.},
    title = {A contribution to the mathematical theory of epidemics},
    year = {1927},
    volume = {115},
    pages = {700-721},
}

@article{Cooper2020,
    doi = {10.1016/j.chaos.2020.110057},
    author = {Ian Cooper and Argha Mondal and Chris G. Antonopoulos},
    journal = {Chaos Solitons Fractals},
    title = {A SIR model assumption for the spread of COVID-19 in different communities},
    year = {2020},
    volume = {139},
}

@article{Gounane2021,
    url = {https://doi.org/10.1515/em-2020-0044},
    title = {An adaptive social distancing SIR model for COVID-19 disease spreading and forecasting},
    author = {Said Gounane and Yassir Barkouch and Abdelghafour Atlas and Mostafa Bendahmane and Fahd Karami and Driss Meskine},
    pages = {20200044},
    volume = {10},
    number = {s1},
    journal = {Epidemiologic Methods},
    doi = {doi:10.1515/em-2020-0044},
    year = {2021},
    lastchecked = {2023-08-08}
}

@article{Marinov2022,
    url = {https://doi.org/10.1038/s41598-022-20276-7},
    title = {Adaptive SIR model with vaccination: simultaneous identification of rates and functions illustrated with COVID-19.},
    author = {Tchavdar T. Marinov and Rossitza S. Marinova},
    pages = {15688},
    volume = {12},
    number = {s1},
    journal = {Nature Scientific Reports},
    doi = {https://doi.org/10.1038/s41598-022-20276-7},
    year = {2022},
}

@article{Perakis2023,
    author = {Perakis, Georgia and Singhvi, Divya and Skali Lami, Omar and Thayaparan, Leann},
    title = {COVID-19: A multiwave SIR-based model for learning waves},
    journal = {Production and Operations Management},
    volume = {32},
    number = {5},
    pages = {1471-1489},
    doi = {https://doi.org/10.1111/poms.13681},
    url = {https://onlinelibrary.wiley.com/doi/abs/10.1111/poms.13681},
    eprint = {https://onlinelibrary.wiley.com/doi/pdf/10.1111/poms.13681},
    year = {2023}
}

@article{Ianni2020,
	author = {Ianni, Aldo and Rossi, Nicola},
	date = {2020/11/04},
	doi = {10.1140/epjp/s13360-020-00895-7},
	id = {Ianni2020},
	isbn = {2190-5444},
	journal = {The European Physical Journal Plus},
	number = {11},
	pages = {885},
	title = {Describing the COVID-19 outbreak during the lockdown: fitting modified SIR models to data},
	url = {https://doi.org/10.1140/epjp/s13360-020-00895-7},
	volume = {135},
	year = {2020}
}

@article{Goel2021,
    doi = {https://doi.org/10.1007/s13278-021-00814-3},
    author = {Rahul Goel and Loïc Bonnetain and Rajesh Sharma and Angelo Furno},
    journal = {Social Network Analysis and Mining},
    title = {Mobility‑based SIR model for complex networks: with case study of COVID‑19},
    year = {2021},
    volume = {11},
    url = {https://doi.org/10.1007/s13278-021-00814-3},
    number = {105},
}

@book{Goodfellow2016,
    title={Deep Learning},
    author={Ian Goodfellow and Yoshua Bengio and Aaron Courville},
    publisher={MIT Press},
    note={\url{http://www.deeplearningbook.org}},
    year={2016}
}

@article{Vortmeyer2021,
    doi = {https://doi.org/10.1038/s41598-021-99609-x},
    author = {Rahel Vortmeyer-Klay and Pascal Nieters and Gordon Pipa},
    journal = {Nature Scientific Reports},
    title = {A trajectory-based loss function to learn missing terms in bifurcating dynamical systems},
    year = {2021},
    volume = {11},
    url = {https://doi.org/10.1038/s41598-021-99609-x},

}

@article{Dandekar2020100145,
    title = {A Machine Learning-Aided Global Diagnostic and Comparative Tool to Assess Effect of Quarantine Control in COVID-19 Spread},
    journal = {Patterns},
    volume = {1},
    number = {9},
    pages = {100145},
    year = {2020},
    issn = {2666-3899},
    doi = {https://doi.org/10.1016/j.patter.2020.100145},
    url = {https://www.sciencedirect.com/science/article/pii/S2666389920301938},
    author = {Raj Dandekar and Chris Rackauckas and George Barbastathis}
}

@misc {Kuwahara2023,
    Title = {Predicting COVID-19 pandemic waves with biologically and behaviorally informed universal differential equations},
    Author = {Kuwahara, Bruce and Bauch, Chris},
    DOI = {10.1101/2023.03.11.23287141},
    Publisher = {medRxiv},
    Year = {2023},
    URL = {https://europepmc.org/article/PPR/PPR632605},
}

@misc{innes2019dont,
    title={Don't Unroll Adjoint: Differentiating SSA-Form Programs}, 
    author={Michael Innes},
    year={2019},
    eprint={1810.07951},
    archivePrefix={arXiv},
    primaryClass={cs.PL},
    URL={https://arxiv.org/abs/1810.07951}
}

@article{bezanson2017,
    author = {Bezanson, Jeff and Edelman, Alan and Karpinski, Stefan and Shah, Viral B.},
    title = {Julia: A Fresh Approach to Numerical Computing},
    journal = {SIAM Review},
    volume = {59},
    number = {1},
    pages = {65-98},
    year = {2017},
    doi = {10.1137/141000671},
    URL = {https://doi.org/10.1137/141000671},
}

@article{Byrne2021,
  doi = {10.21105/jcon.00068},
  url = {https://doi.org/10.21105/jcon.00068},
  year = {2021},
  publisher = {The Open Journal},
  volume = {1},
  number = {1},
  pages = {68},
  author = {Simon Byrne and Lucas C. Wilcox and Valentin Churavy},
  title = {MPI.jl: Julia bindings for the Message Passing Interface},
  journal = {Proceedings of the JuliaCon Conferences}
}

@article{Nogueira2022,
  doi = {https://doi.org/10.1002/cjce.24495},
  year = {2022},
  volume = {100},
  number = {9},
  pages = {2279-2290},
  author = {Idelfonso B. R. Nogueira and Vinicius V. Santana and Ana M. Ribeiro and Alirio E. Rodrigues},
  title = {Using scientific machine learning to develop universal differential equation for multicomponent adsorption separation systems},
  journal = {The Canadian Jornal of Chemical Engineering}
}
\end{document}